\documentclass[sigconf]{acmart}

\usepackage[normalem]{ulem}
\usepackage{diagbox}
\AtBeginDocument{%
  \providecommand\BibTeX{{%
    \normalfont B\kern-0.5em{\scshape i\kern-0.25em b}\kern-0.8em\TeX}}}

\copyrightyear{2024} 
\acmYear{2024} 
\setcopyright{acmcopyright} 

\acmConference[SIGIR '24]{Proceedings of the 47th International ACM SIGIR Conference on Research and Development in Information Retrieval}{July 14--18, 2024}{Washington, D.C., USA}

\acmISBN{978-1-4503-XXXX-X/18/06}

\usepackage{booktabs} 
\usepackage{color}
\usepackage{graphicx}
\usepackage{url}

\begin{document}

\title{Surprising Efficacy of Fine-Tuned Transformers for Fact-Checking over Larger Language Models} 


\author{Vinay Setty}
\affiliation{%
  \institution{Factiverse AI}
  \country{Norway}
}
\email{vinay@factiverse.ai}

\begin{abstract}
In this paper, we explore the challenges associated with establishing an end-to-end fact-checking pipeline in a real-world context, covering over 90 languages. Our real-world experimental benchmarks demonstrate that fine-tuning Transformer models specifically for fact-checking tasks, such as claim detection and veracity prediction, provide superior performance over large language models (LLMs) like GPT-4, GPT-3.5-Turbo, and Mistral-7b. However, we illustrate that LLMs excel in generative tasks such as question decomposition for evidence retrieval. Through extensive evaluation, we show the efficacy of fine-tuned models for fact-checking in a multilingual setting and complex claims that include numerical quantities.

\end{abstract}

\ccsdesc[500]{Information systems~Retrieval tasks and goals}

\keywords{Multilingual; Fact-checking}

\maketitle

\section{Introduction}

The spread of online misinformation poses a significant challenge globally, impacting societies, political landscapes, and public opinions. This problem is compounded by the complexity of fact-checking, a task that is challenging even for humans\footnote{\url{https://www.niemanlab.org/2024/01/asking-people-to-do-the-research-on-fake-news-stories-makes-them-seem-more-believable-not-less/}}. Research communities specializing in natural language processing and information retrieval have made significant strides in automating aspects of this task, addressing some challenges.

However, several challenges remain that hinder the widespread adoption of these technologies in the industry. These challenges include but not limited to multilinguality, verifying numerical claims, and the  temporal aspects of facts~\cite{Guo:2022:JACL,Opdahl:2023:DKE}. The emergence of large language models (LLMs), such as GPT-4, Gemini and Mistral, has raised expectations that they might address these issues effectively. There are many recent works that propose use of LLMs for end-to-end fact-checking pipeline~\cite{Wang:2023:arXiv,Chern:2023:arXiva}. Particularly, LLMs, being inherently multilingual, are anticipated to excel in processing and understanding non-English languages, especially those considered low-resource. Furthermore, advanced models like GPT-4 have shown increased capabilities in numerical reasoning, which could potentially enhance the accuracy of fact-checking numerical data.

Despite expectations, this paper demonstrates that, for certain fact-checking tasks like identifying check-worthy claims and assessing their veracity, specialized Transformer models perform better than LLMs in practical situations. Nonetheless, it's worth mentioning that LLMs are superior in generative tasks, such as decomposing claims into queries and posing questions, which helps in the more effective gathering of evidence. This paper specifically addresses fact-checking within multilingual contexts and for claims involving numerical quantities. We show that multilingual Transformer models, such as XLM-RoBERTa, fine-tuned for select languages, can surpass LLMs in effectiveness across more than 90 languages. Similarly, Transformer models specialized in numerical question answering demonstrate superior performance.

An additional obstacle encountered by fact-checking tools in the industry pertains to privacy concerns. Journalists and fact-checkers, the primary users of these technologies, hesitate to transmit sensitive and private information to third-party servers that host LLMs. This paper explores the effectiveness of fine-tuned models and smaller, self-hostable LLMs in diverse fact-checking situations.
\begin{figure*}[h!!]
    \includegraphics[width=\textwidth]{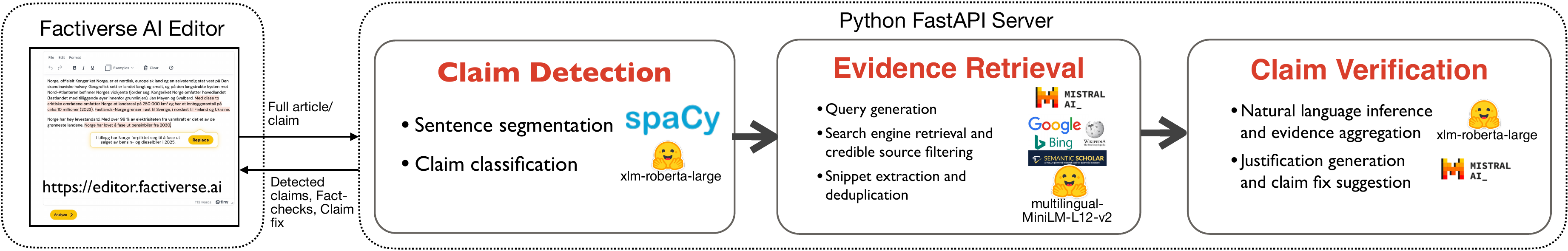}
    \caption{System Architecture of the Fact-Checking Pipeline at Factiverse}
    \label{fig:system_arch}
\end{figure*}
\subsection{Related Work}
Recently, the use of machine learning models for fact-checking has gained attention in the academic research, yet its implementation in the industry remains limited. Factiverse has a few commercial tools capable of fact-checking textual content\footnote{\url{https://editor.factiverse.ai}}~\cite{Botnevik:2020:SIGIRa,Mishra:2019:ICTIR}. Meanwhile, large language model (LLM) chat applications like ChatGPT and Gemini have started providing references with their outputs. Similarly, there are writer tools such as Originality AI\footnote{\url{https://originality.ai}} which use LLMs to do the fact-checking.  However, these solutions still can still generate erroneous text, and they rely on a limited range of sources for their references. Factiverse has also released a custom GPT to fact-check ChatGPT generated text\footnote{\url{https://gpt.factiverse.ai}}. There are also tools such as Google Search-Augmented Factuality Evaluator (SAFE)~\cite{Wei:2024:arXiv}, SelfCheckGPT~\cite{Manakul:2023:arXiv} FactTool~\cite{Wang:2023:arXiv} and FAVA~\cite{Chern:2023:arXiva} to identify factual inaccuracies in LLM outputs using search results and prompts.  Despite their complexity, these tools have not been adapted for real-world applications.

\section{Fact-checking pipeline}
\label{sec:method}

We follow a three-stage fact-checking pipeline (shown in Figure \ref{fig:system_arch}) that is widely used in the literature~\cite{Guo:2022:JACL}. We have a front-end React app, which can be used to compose articles and fact-check them\footnote{\url{https://editor.factiverse.ai}}. The frontend communicates with the backend via the REST APIs.

The first stage identifies check-worthy sentences (also referred to as `claims' in the rest of the paper), in the given text, either written by users or generated using LLMs. The second stage gathers evidence from the open web and previous fact-checks for a given claim. Finally, the claim is verified using the gathered evidence in the third stage. The pipeline is hosted using a Python FastAPI backend that is hosted on a Kubernetes cluster with auto-scaling on the Google Cloud Platform. There are several ML models used in the backend targeted for (a) Check-worthy claim detection, (b) Evidence search, and (c) Veracity prediction. We now explain the sub-tasks within these steps of the pipeline.

\subsection{Check-worthy Claim Detection}
\label{sec:claim_detection}
The objective of this stage is to enrich the  sentences within the text that merit fact-checking. We first segment sentences using spaCy models. While there is no fixed definition, it is generally agreed upon which sentences require verification: (1) they invite the public to scrutinize their accuracy and authenticity, and (2) they exclude subjective statements such as opinions, beliefs, or queries \cite{Panchendrarajan:2024:arXiv}.

We fine-tune the \textit{XLM-RoBERTa-Large} model \footnote{\url{https://huggingface.co/FacebookAI/xlm-roberta-large}}~\cite{DBLP:journals/corr/abs-1911-02116}, using the data from ClaimBuster~\cite{Hassan:2017:VLDB} and CLEF CheckThat Lab!~\cite{Alam:2021:arXiv} along with a dataset collected from Factiverse production system (see Table \ref{tab:dataset}) to classify sentences into `Check-worthy' and `Not check-worthy'.

\subsection{Evidence Search}
The objective of the second stage is to retrieve highly relevant articles essential for fact-checking the verification of the check-worthy claims detected in the previous phase. 
Since using a claim verbatim as a search query may not yield sufficiently relevant results, we need to decompose it into questions to diversify the search results.
In this regard, we employ Large Language Models (LLMs), to generate targeted questions to verify the claim. The prompt used for this purpose can be found in\footnote{\url{https://github.com/vinaysetty/factcheck-editor/blob/main/code/prompts/prompts.py}}.
We then leverage a diverse range of search engines including Google, Bing, You.com, Wikipedia, Semantic Scholar\footnote{\url{https://www.semanticscholar.org}} (contains 212M scholarly articles), and Factiverse's fact-checking collection\footnote{\url{https://factisearch.ai}}, which comprises 280K fact-checks updated in real-time.
We then deduplicate the search results by matching URLs, titles, and content with approximate similarity. Furthermore, to enhance the relevancy of the data, we select the top three paragraphs that are most similar to the claim using a multilingual sentence encoder~\cite{Reimers:2019:Arxiv}.

\begin{figure*}[t!!]
    \includegraphics[width=\linewidth]{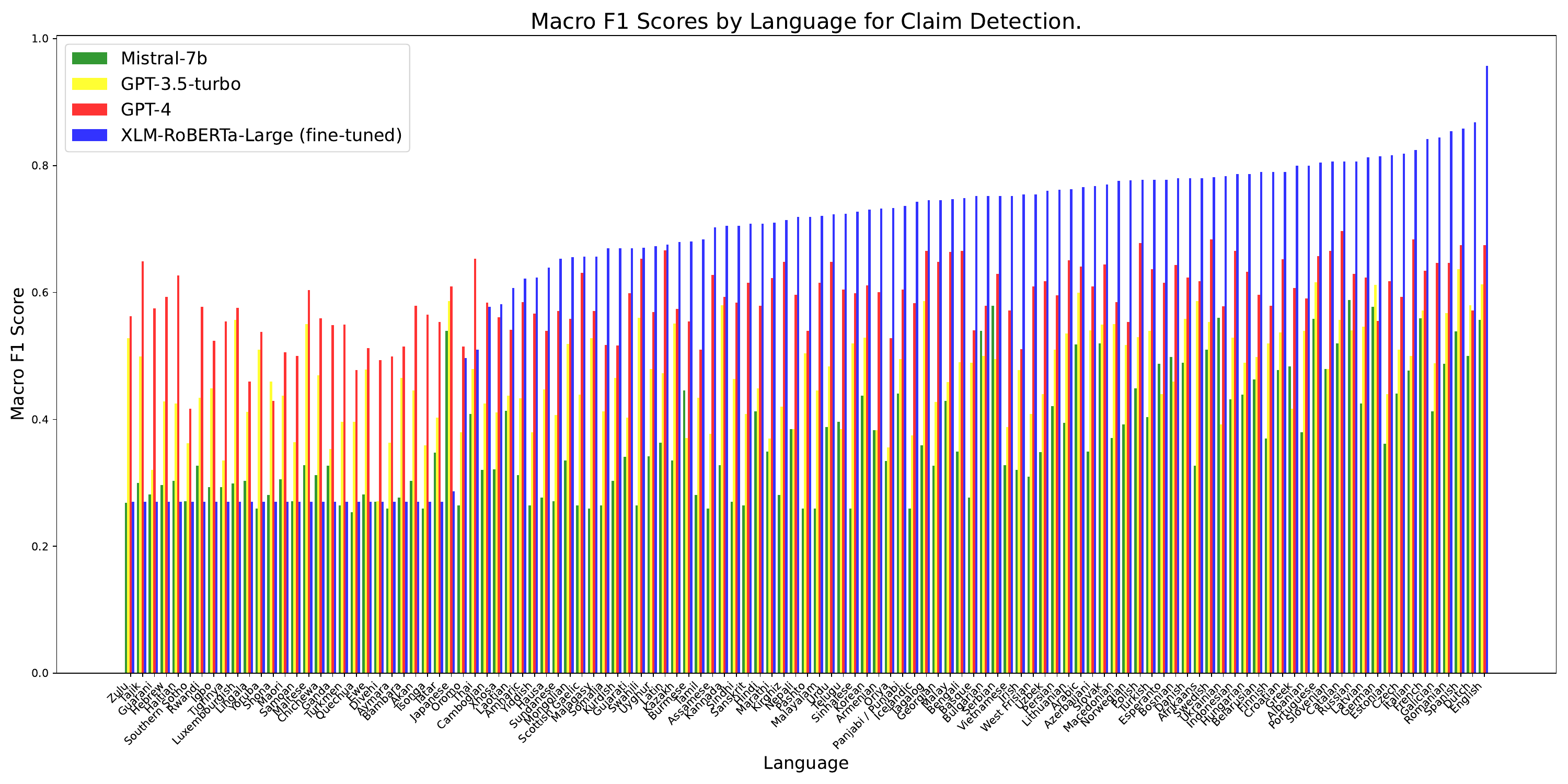}
    \caption{Evaluation of claim detection for 114 languages using Factiverse model, GPT-3.5-Turbo, GPT-4 and Mistral-7b.}
    \label{fig:claim_detection_macro}
\end{figure*}

\subsection{Veracity Prediction}
The final step in the fact-checking pipeline is predicting the veracity of the claim based on the evidence. 
To ensure the evidence is coming from credible sources, we exclude websites known to spread misinformation\footnote{\url{https://en.wikipedia.org/wiki/List_of_fake_news_websites}}. 
We first predict the stance of the evidence for the claim (natural language inference or NLI). I.e, if the claim is supported or refuted by the evidence. 
To simplify, we omit claims which are half true or false, similar to existing works~\cite{Schlichtkrull:2023:arXiva}. 
For this step, we also fine-tune \textit{XLM-RoBERTa-Large} model using a combination of FEVER data and real-world fact-checks from Factisearch.
We then aggregate the stance predictions for individual evidence snippets by majority voting following existing works~\cite{Popat:2017:WWWa}. 
However, we acknowledge that there is no consensus on how to aggregate multiple evidence snippets and account for the credibility of these sources.
We further summarize the evidence snippets,  using an LLM, which acts as a justification for the veracity prediction. In addition, we also generate a correction for the refuted claims based on the justification summary using an LLM.

\section{Experimental Setup}
\label{sec:eval}

\paragraph{Dataset:} We use a dataset sample from production deployment at Factiverse\footnote{\url{https://github.com/vinaysetty/factcheck-editor/tree/main/data}}. Since the original data was only in English, we translated the claims into 114 languages using the Google Translate API. An overview of the dataset is shown in Table \ref{tab:dataset}. Contrary to existing fact-checking datasets from fact-checking organizations, this sample has `True' class, the majority class. This is because Factiverse applications focus on fact-checking full articles, while manual fact-checkers tend to pick `False' claims to verify. For numerical claims, we sample 100 claims from a collection of past fact-checks from Factisearch\footnote{\url{https://factisearch.ai}} with numerical quantities\footnote{\url{https://github.com/vinaysetty/factcheck-editor/tree/main/data/veracity_prediction}}.

\paragraph{Models:} 
For claim detection and veracity prediction tasks, we fine-tune XLM-RoBERTa-Large~\cite{DBLP:journals/corr/abs-1911-02116}. For verifying numerical claims, we also use FinQA-RoBERTa-Large~\cite{Zhang:2022:NEURIPS} fine-tuned using the financial QA dataset, FinQA~\cite{Chen:2021:EMNLP}. We use GPT-4 and GPT-3.5-Turbo and Mistral-7b~\footnote{\url{https://huggingface.co/mistralai/Mistral-7B-Instruct-v0.2}}  as LLMs. To adapt LLMs to perform fact-checking, we do prompt engineering to draft a prompt to predict both check-worthiness of a claim and for veracity prediction. To make the comparison fair, the same prompts are used for all LLMs\footnote{Prompts can be found here: \url{https://github.com/factiverse/factcheck-editor/blob/main/code/prompts/prompts.py}}. We set the temperature to `0.2', set a random seed and repeat all experiments 3 times and report the mean scores.

For question decomposition, in addition to LLMs, we also compare with T5-3b~\footnote{\url{https://huggingface.co/google-t5/t5-3b}}\cite{Raffel:2020:JLMR} fine-tuned using the ClaimDecomp dataset~\cite{Chen:2022:arXiva}. Since T5-3b model only supports English, we perform question decomposition effectiveness analysis only for English claims. 
    
\begin{table}[t!!]
    \caption{Dataset distribution.}
    \centering
    \begin{tabular}{l|rrrrr}
    \hline
      \textbf{Split}   & \textbf{Not Check-} & \textbf{Check-} & \textbf{True} & \textbf{False} &\textbf{Total} \\
         & \textbf{worthy} & \textbf{worthy }& \textbf{Claims} & \textbf{Claims} & \\
      \midrule
       Train  & 609 & 548 & 332 & 196 & 1,076  \\
       Dev  & 38 & 25 & 15 & 10 & 63  \\
       Test  & 62 & 38 & 26 & 12 & 100   \\
       \hline
    \end{tabular}
    \label{tab:dataset}
\end{table}

\begin{table*}[h!!]
\centering
\caption{Comparison of Effectiveness of Question Decomposition Methods for Veracity Prediction for English claims. Rows contain question decomposition methods and columns contain natural language inference (NLI) methods.}
\label{tab:question_generation}
\begin{tabular}{l|cc|cc|cc|cc}
\hline
\diagbox{\textbf{Question}}{\textbf{NLI}} & \multicolumn{2}{c|}{\textbf{XLM-RoBERTa-Large}} & \multicolumn{2}{c|}{\textbf{Mistral-7b}} & \multicolumn{2}{c|}{\textbf{GPT-3.5-Turbo}} & \multicolumn{2}{c}{\textbf{GPT-4}} \\
\textbf{Decomposition} & \textbf{Macro F1} & \textbf{Micro F1} & \textbf{Macro F1} & \textbf{Micro F1} & \textbf{Macro F1} & \textbf{Micro F1} & \textbf{Macro F1} & \textbf{Micro F1} \\
\hline
T5-3b (fine-tuned)          & 0.632             & 0.639             & 0.398                & 0.444                & 0.438               & 0.444               & 0.469               & 0.473               \\
Mistral-7b      & 0.642             & 0.647             & 0.410                & 0.412                & 0.454               & 0.471               & 0.463               & 0.470               \\

GPT-4           & 0.719             & 0.722             & 0.550                & 0.556                & 0.494               & 0.500               & 0.494               & 0.500               \\
GPT-3.5-Turbo    & \textbf{0.741 }            & \textbf{0.750}             & \textbf{0.556}                & \textbf{0.562}                & \textbf{0.498 }              & \textbf{0.500}               & \textbf{0.530}              & \textbf{0.531}               \\
\hline
\end{tabular}
\end{table*}

\paragraph{Metrics:} We use the Macro-F1 and Micro-F1 scores to compare the performance of the models since there is an imbalance in the classes (applies to both claim detection and veracity prediction tasks). For multilingual evaluation, we only show Macro-F1 but a similar trend was observed in Micro-F1.

The code used for the experiments is available on GitHub\footnote{\url{https://github.com/factiverse/factcheck-editor}}.

\section{Experimental Results}

\subsection{Claim Detection}
As shown in Figure \ref{fig:claim_detection_macro}, the fine-tuned XLM-RoBERTa-Large outperforms both OpenAI and Mistral models in most languages. Since the model was trained mostly in English, it is unsurprisingly the best-performing language. For some languages (towards the left side of the plot), we see that LLMs perform better. On closer inspection, these are the languages not yet supported by the model. Mistral-7b seems to be the worst-performing model overall. It seems to be because Mistral struggles to follow instructions in the prompt for text classification and hallucinates more than other models. Table \ref{tab:result_summary} shows the average Macro-F1 and Micro-
F1 scores for all four models. This suggests that a fine-tuned model is significantly better for claim detection compared to using LLMs with prompt engineering.

\begin{table}
    \caption{Claim detection and veracity prediction results presented as mean Micro and Macro-F1 scores for all languages.}
    \centering
    \begin{tabular}{l|rr|rr}
    \hline
      \textbf{Model}   & \multicolumn{2}{c}{\textbf{Claim Detection}}  & \multicolumn{2}{c}{\textbf{Veracity Prediction}}  \\
         & \textbf{Ma.-F1} & \textbf{Mi.-F1 }& \textbf{Ma.-F1} & \textbf{Mi.-F1 } \\
      \midrule
      Mistral-7b  & 0.477 & 0.510 & 0.509 & 0.557  \\
      
       GPT-3.5-Turbo  & 0.562 & 0.567 &  0.440 & 0.396  \\
      GPT-4  & 0.624 & 0.591 &  0.460 & 0.426  \\ 
       XLM-RoBERTa-Large  & \textbf{0.743} & \textbf{0.768} & \textbf{0.575} & \textbf{0.594}   \\
       \hline
    \end{tabular}
    \label{tab:result_summary}
\end{table}
\subsection{Evidence Retrieval}
Here, we compare the effectiveness of different question decomposition methods for evidence retrieval. In this paper, we report the effect of these methods only on the final task of veracity prediction. We leave the evaluation of question quality and information retrieval measures for a future work.

Table \ref{tab:question_generation} shows that GPT-3.5-Turbo outperforms other question decomposition methods, including GPT-4, with the fine-tuned T5-3b model ranking lowest. This suggests LLMs are better at generating questions for evidence retrieval, though GPT-4's longer questions may cause topic drift in search results.

\begin{table}[t!!]
\centering
\caption{Effectiveness of fine-tuned models compared to LLMs for numerical claims.}
\label{tab:numclaims}
\begin{tabular}{l|cc|cc}
\hline
\diagbox{\textbf{Question}}{\textbf{NLI}} & \multicolumn{2}{c|}{\textbf{XLM-RoBERTa}} & \multicolumn{2}{c}{\textbf{FinQA-RoBERTa}}  \\
\textbf{Decomposition} & \textbf{Ma. F1} & \textbf{Mi. F1} & \textbf{Ma. F1} & \textbf{Mi. F1}   \\
\hline

T5-3B (fine-tuned) &   0.579 & 0.611 & 0.642            & 0.750                    \\

GPT-4   & 0.642 &  0.700      & 0.641             & 0.750              \\
GPT-3.5-Turbo  & 0.562            & 0.737   & 0.733& 0.755                \\
Mistral-7b  &   \textbf{0.672 }& \textbf{0.722} & \textbf{0.781}            & \textbf{0.842}    \\
\hline
\end{tabular}
\end{table}

The fine-tuned XLM-RoBERTa-Large NLI model consistently outperforms others across all question generation methods. In contrast, the fine-tuned T5-3b emerges as the least effective question generation method. For this particular task, Large Language Models (LLMs) exhibit markedly superior performance compared to the fine-tuned T5-3b model. Furthermore, it appears that the larger the LLM, the better the performance. While self-hosted Mistral-7b does not seem to generate effective questions for evidence retrieval, it is worth exploring other open larger models in future.
\begin{figure}[t!!!]
    \includegraphics[width=\linewidth]{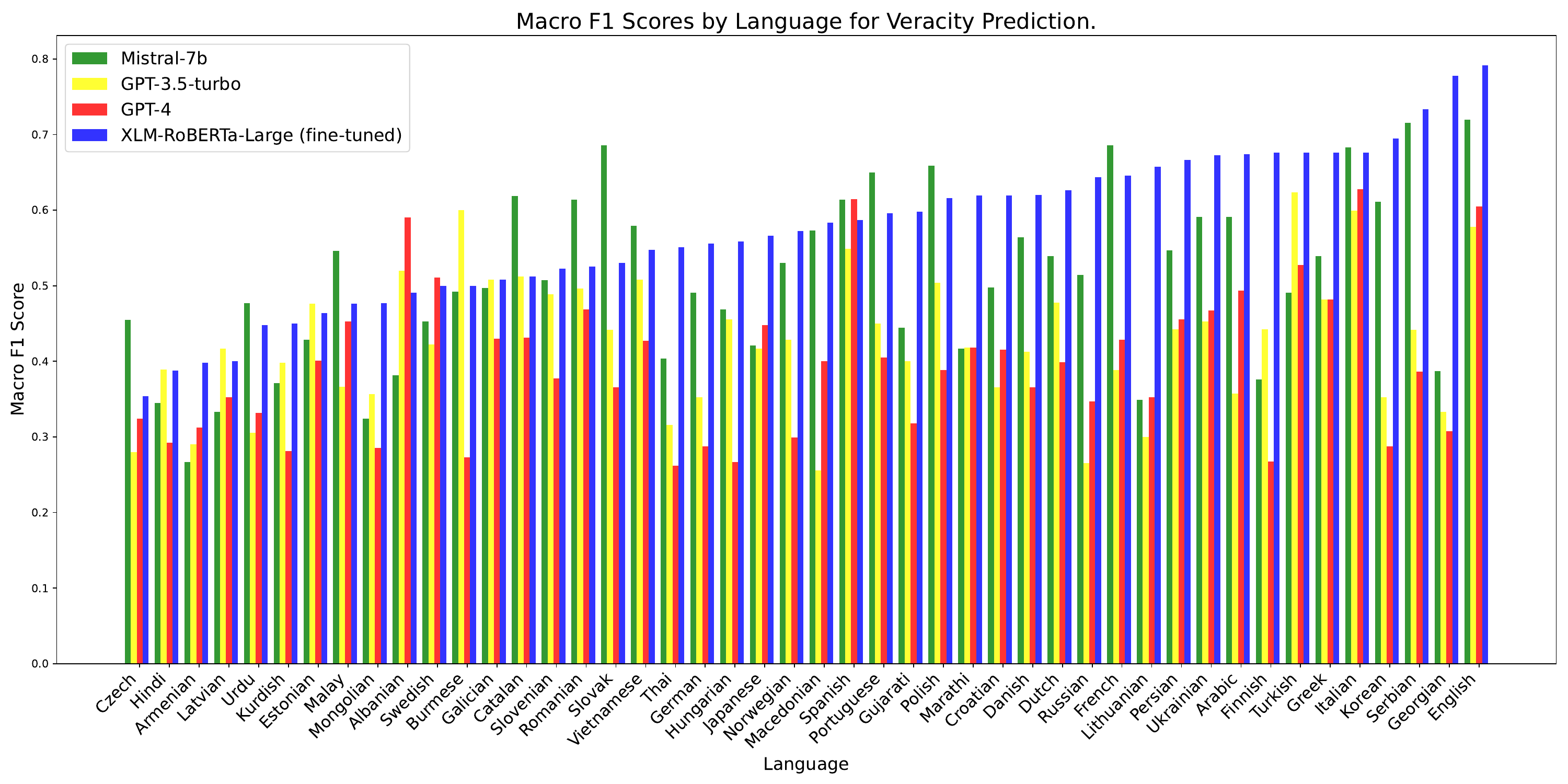}
    \caption{Evaluation of veracity prediction for 46 languages.}
    \label{fig:nli_eval}
\end{figure}
\subsection{Veracity Prediction}
As shown in the Figure, \ref{fig:nli_eval}, the fine-tuned XLM-RoBERTa-Large model outperforms the other models in 37 languages. GPT-4 is the best model only for three languages: Swedish, Albanian, and Georgian. Mistral-7b is the best model in 8 languages, and it is interesting to see that Mistral performs better than OpenAI models despite being a much smaller LLM. Unsurprisingly, Mistral-7b seems to be the best model for some European languages, such as French, Spanish, Catalan, and Portuguese. Due to the lack of evidence snippets for some languages, they were excluded. Veracity prediction results are summarized in Table \ref{tab:result_summary}.

Table \ref{tab:numclaims} compares the fine-tuned XLM-RoBERTa-Large and the FinQA-RoBERTa-Large for numerical claims fine-tuned by \cite{V:2024:arXiv}. Not surprisingly, FinQA-RoBERTa-Large performs better because it is tailored to perform reasoning over numerical data~\cite{Zhang:2022:NEURIPS}. However, it is surprising to note that the question decomposition by Mistral-7b is better for numerical claims than OpenAI and T5 models. We believe this is because the numerical claims need different style of questions, which Mistral seems to perform better. We omit the comparison of LLMs for the NLI task due to lack of space, but the performance is similar to the results in Table \ref{tab:question_generation}.

\section{Conclusion}
In this paper, we explored the complexities involved in fact-checking systems and demonstrated that fine-tuned, smaller models surpass LLMs in multilingual contexts and when evaluating numerical claims. However, LLMs exhibit superior performance in generative tasks such as question decomposition. The outcomes suggest that both fine-tuning Transformer models and employing self-hosted LLMs could be effective strategies for enhancing fact-checking operations. Given the preliminary nature of this research and smaller scale of the dataset used, it becomes imperative to undertake further studies to confirm these findings. Future work will delve into evaluating the evidence retrieval step and larger scale experiments.

\section{Acknowledgements}
This work is in part funded by the Research Council of Norway project EXPLAIN (grant number 337133) with partners from Factiverse and TU Delft.

\newpage
\bibliographystyle{ACM-Reference-Format}
\balance
\bibliography{sigir2024-ai-editor.bib}

\end{document}